\documentclass[10pt,twocolumn,letterpaper]{article}

\usepackage{iccv}
\usepackage{times}
\usepackage{epsfig}
\usepackage{graphicx}
\usepackage{amsmath}
\usepackage{amssymb}
\usepackage{subfigure}
\usepackage{algorithm}
\usepackage{algorithmic}
\usepackage{tabularx}
\usepackage{verbatim} 
\usepackage{mathtools} 
\usepackage{varwidth} 
\usepackage{rotating}
\usepackage{tabularx}
\usepackage[table]{xcolor}
\usepackage{color, colortbl}
\usepackage{array,booktabs,arydshln,xcolor}
\usepackage{multirow}

\usepackage[pagebackref=true,breaklinks=true,letterpaper=true,colorlinks,bookmarks=false]{hyperref}

\iccvfinalcopy 

\ificcvfinal\fi
\begin{document}

\title{Large Scale Visual Recommendations From Street Fashion Images}

\author{
Vignesh Jagadeesh, Robinson Piramuthu, Anurag Bhardwaj, Wei Di, Neel Sundaresan\\
eBay Research Labs, 2065 East Hamilton Avenue, San Jose, CA-95128\\
{\tt\small \{vjagadeesh, rpiramuthu, anbhardwaj, wedi, nsundaresan\}@ebay.com}
}

\maketitle

\begin{abstract}
We describe a completely automated large scale visual recommendation system for fashion. Our focus is to efficiently harness the availability of large quantities of online fashion images and their rich meta-data. Specifically, we propose four data driven models in the form of Complementary Nearest Neighbor Consensus, Gaussian Mixture Models, Texture Agnostic Retrieval and Markov Chain LDA for solving this problem. We analyze relative merits and pitfalls of these algorithms through extensive experimentation on a large-scale data set and baseline them against existing ideas from color science. We also illustrate key fashion insights learned through these experiments and show how they can be employed to design better recommendation systems. Finally, we also outline a large-scale annotated data set of fashion images (\textbf{Fashion-136K}) that can be exploited for future vision research.
\end{abstract}

\section{Introduction}
\label{sec:Introduction}


Availability of huge amounts of online image data today has immense potential to create novel interaction mechanisms that go beyond text to other modalities such as images. Most progress in computer vision in the past decade has been partly because of annotated public data sets. Obtaining large data set with reasonably clean annotations is a big challenge. Much of the available data sets have been focusing on aspects such as scenes \cite{TorralbaSun10}, generic categories \cite{FeiFeiImageNet09}, etc.

\begin{figure}[ht!]
     \begin{center}
            \includegraphics[width=0.8\linewidth]{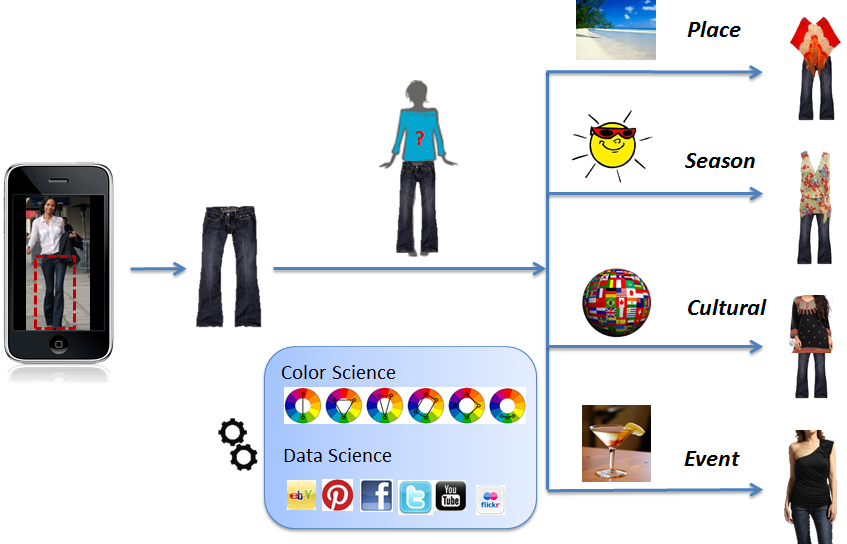}
    \end{center}
    \caption{Illustration of context based recommendation. Different tops are recommended for the same pair of blue jeans.}
   \label{fig:themeQuery}
   \vspace{-2.5ex}
\end{figure}

Our focus is on street fashion, where visual experiences have strong social, cultural and commercial importance. Street fashion images typically contain full view of a street fashion model striking a pose with several fashionable accessories. The first large scale public data set for street fashion was presented in \cite{YamaguchiCVPR12}. It has 158235 images, but only 685 images were annotated. The authors present interesting problems such as parsing clothing, improving pose estimation using clothing prediction and garment retrieval independent of pose.

\begin{figure*}[t]
\centering
\vspace{-2ex}

\begin{tabular}{ccc}
\raisebox{1.4cm}
{
\parbox{.24\linewidth}{
\footnotesize{
\begin{tabular}{ll}
\hline\hline \\ [-2ex]
\textbf{Item} & \textbf{Quantity} \\
\hline \\ [-2ex]
Initial size of corpus      & 196974\\
Final size after cleanup        & 135893\\
($H\ge400, H/W>1$) & \\
Number of unique tags      & 71\\
Number of unique users     & 8357\\
Number of unique brands     & 20110\\
Number of known cities      & 115\\
Number of known regions & 136\\
[1ex]
\hline
\end{tabular}
}
}
}&
\includegraphics[width=0.1\linewidth]{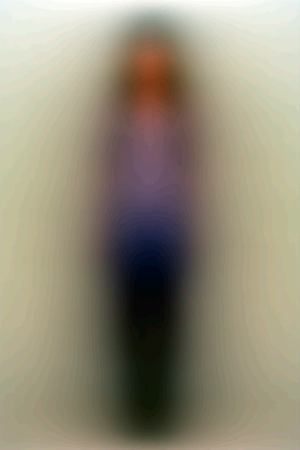} &
\includegraphics[width=0.57\linewidth]{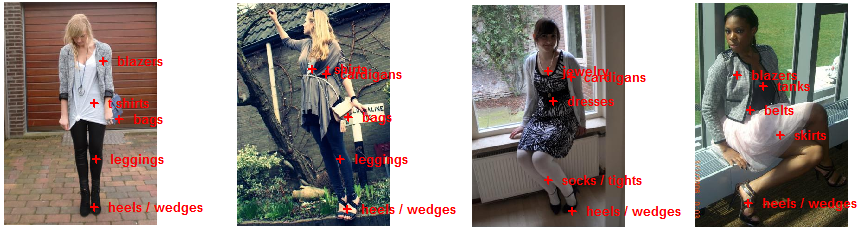}
\\
\footnotesize{\hspace{0\linewidth}(a) Images, tags, geo-location.} &
\footnotesize{\hspace{0\linewidth}(b) Average image.}&
\footnotesize{\hspace{0\linewidth}(c) Exemplars showing various pose and complex background.} \\
\end{tabular}
\vspace{0em}
\caption{Basic summary of street fashion corpus Fashion-136K. Images were posted and tagged by fashion designers and fashionistas. There are about 3-4 annotations per image, with each user posting anywhere from 8 to 524 images. Most contain full view of model.}
\label{fig:AvgImageCorpus}
\vspace{-3ex}
\end{figure*}

We gathered a street fashion data set, which we call Fashion-136K that
contains street fashion images along with other meta data such as
annotations of accessories, brands and demographics of fashionistas
who posted those images. Such a rich source of information has never
been exploited before in learning fashion concepts. In this work, we
focus on the problem of learning complementary clothing recommender
systems using the Fashion-136K data set. A brief
illustration of the data is shown in Figure~\ref{fig:AvgImageCorpus}.

One scenario that can be envisioned in the context of fashion is ``up-selling in e-commerce" where an online shopper with an item in her shopping cart is recommended other items that complement the cart. This paper proposes a similar application for fashion, where given an image of a fashion item (say ``jeans"), the goal is to recommend matching fashion items (say ``tops") that complement the given item (Figure~\ref{fig:themeQuery}).  

The computational challenges involved in the problem are two fold. Firstly, the representation of fashion item as visual features is an open problem. This problem is compounded by the task of learning inter-dependencies between such features from different items (i.e. ``tops" and ``jeans"). Further, a number of practical challenges that need to be addressed include obtaining high quality image data with clean background, scalability issues in terms of memory and speed requirements that can handle large image databases as well as generate real-time predictions. Such constraints are even more crucial in the context of mobile shopping which is limited by the compute, memory and network capacities. Finally, the notion of relevant matches may depend on a number of factors such as location, season as well as occasion which makes the task of model learning inherently complex.
 
In this paper, we propose to use data-driven approaches to address the
aforementioned issues. We formulate the recommendation problem as
follows:  Given an image $i$ containing a set of fashion items, also
referred to as ``parts" (i.e. sunglasses, tops, skirts, shoes,
handbags), the holistic description $\underline{H}_i$ of an image $i$
is given by $\underline{H}_i^T := [\underline{h}_{i1}^T,
\underline{h}_{i1}^T,\ldots, \underline{h}_{iP}^T] \in \Re^{PK},
\underline{h}_{ij}^T\mathbf{1}_K=1, \forall j \in [1, 2, . . P]$,
where $P$ is the number of parts, $K$ is the size of common code book to
represent parts and $\underline{h}_{ij} \in \Re^{K}$ denotes the
representation of part $j$ for image $i$. Our task is to learn a
predictive model $\mathcal{M}(\mathbf{H},\mathbf{q})$ where $\mathbf{H} =
[\underline{H}_1, \underline{H}_2,\ldots,\underline{H}_n]^T$ for $n$
number of images in the data set. Given an input query vector $\underline{H}_q^T =  [\underline{h}_{q1}^T, \underline{h}_{q2}^T. . .\underline{h}_{qm}^T,\ldots,\underline{h}_{qP}^T] \in \Re^{PK}$, with a
missing column entry corresponding to part $m$, the problem of model learning reduces to predicting the value of empty column in the vector using Model $\mathcal{M}$. Once the model is learned, it can be used to transform the input query which can then be used in applications such as recommendation systems, retrieval and personalization.

\section{Related Work}

There have been only handful attempts to solve the problem of visual fashion recommendation. We are aware of only two existing works in this area. Iwata \etal~\cite{IwataIJCAI11} study this problem in isolation where they propose a topic model based approach to solve this problem. However, given the small data set size of their experimentation it is difficult to ascertain the relative merits of their system. Liu \etal~\cite{LiuMM12}
propose a latent SVM based formulation to handle both ``wear properly" and ``wear aesthetically" criteria in their model. However most of their experimentation is tailored towards solid colored clothing and their qualitative analysis fails to demonstrate the efficacy of system performance on a variety of clothing patterns.
 
Other related papers in visual fashion~\cite{ChenECCV12, LiuCVPR12,
  weber2011part, YamaguchiCVPR12,yang2011real} work on the problem of
fashion parsing and similarity retrieval where the goal is to retrieve
similar fashion images for a given query. These methods either employ
the mixture of parts based pose estimator or use the poselets based
part detectors after applying the deformable parts based detector for
person detection. Yamaguchi \etal\cite{YamaguchiCVPR12} utilize a
superpixel labeling approach on a CRF for inferring part labels from a
labeled data set of fashion apparel. The Street to Shop
system~\cite{LiuCVPR12} attempts to solve the cross domain discrepancy
between catalog images, and query images captured in the
wild. Gallagher \etal~\cite{ChenECCV12} utilize attribute based
classifiers regularized by a CRF that model interactions between
attributes. We believe such methods are limited by their ability to
accurately parse humans from real-world images in the wild since the problem of person detection and subsequent pose estimation is still largely unsolved~\cite{RamananPose11}.

There exists a vast amount of literature on learning aesthetically pleasing color contexts which can provide a great insight for a deeper understanding of this problem. In these papers~\cite{cohen2006color, li2009aesthetic} the goal is to present the user with matches that are perceptually pleasing along with the query item. As a result of this research, the Matsuda templates have emerged as a popular choice for representing color basis functions. This allows for a straightforward retrieval mechanism by rotating the color wheel according to some preconceived rule. For instance, complementary color retrieval simply shifts the hue wheel by 180 degrees and subsequently retrieves nearest neighbors. In this paper, we propose to use these techniques as baselines and compare their performance against multiple data driven approaches.


\section{Proposed Methods for Recommendation}

Since very limited research exists for investigating the full spectrum for solving the problem of visual recommendation, we propose the following set of algorithms to address this issue. This also allows to exploit the relative merits of each algorithm for designing a better solution. We split a data set of $n$ images into $n_{train}$ training and $n_{test}$
testing images where each image has a set
of parts, such as head, foot, torso, etc. Let us assume part indices
to be represented by $p=\{1, 2, 3,\ldots, P\}$. For example, $p=1,2,P$
could correspond to head, torso and foot, respectively. Some of these parts
are observable to the algorithm, while the rest are hidden. The aim of
the algorithm is to infer features related to the hidden parts. On a
similar note, let us assume visible part indices to be $p_v = \{1, 2,
... , |p_v|\}$ and hidden part indices to be
$p_h=\{1,2,...|p_h|\}$. Note that $|p_v|+|p_h|=P$.

We assume each part $j$ of image $i$ to have an associated part descriptor $\underline{h}_{ij}\in\Re^K$. The
feature $\underline{h}_{ij}$ can be color, texture, or any other visual descriptor such as bag of words (BoW). An
$i^{th}$ image can now be described by the concatenation of all part
descriptors:
\begin{equation}
\underline{H}_i^T := [\underline{h}_{i1}^T, \underline{h}_{i2}^T,\ldots, \underline{h}_{iP}^T]
\end{equation}

In case of a $K$-dimensional HSV histogram, $K$ is the number of bins, while in case of bag of words, $K$ is the size of code book. Summarizing the above notation, given a database of $n$ images $I_1,I_2 ... I_n$, with representations $\underline{H}_1^T, \underline{H}_2^T, . . . \underline{H}_n^T$ divided into $n_{train}$ and $n_{test}$ training and validation sets, without loss of generality, we can re-order images and parts such that the first  $n_{train}$ are images are for training and the last (i.e. $P^{th}$) part is hidden. This means $p_v = \{1,2, ... , P-1 \}$ and $p_h = \{P\}$,
and $|p_v| = P-1$ and $|p_h| = 1$. We wish to reiterate that $p_v$ and $p_h$ may be allowed to change, while the formulations below still hold.

\subsection{Tuned Perceptual Retrieval (PR)}
This baseline uses the popular perceptually motivated Matsuda
templates \cite{cohen2006color} and the transformation is either a
complementary or triad pattern for the hue histogram. The saturation
and value channels are simply reflected on the intuition that people
prefer wearing contrasting styles.

\subsection{Complementary Nearest Neighbor Consensus (CNNC)}
Let us define a fixed metric on the space, say $dist$. The metric
could be $||.||_1, ||.||_2,$ KL-divergence, or Earth Mover's
distance. We stick to $||.||_1$ for simplicity. The problem we are faced with is predicting the hidden $P^{th}$
part. Denoting a single test query sample as $q$,
\begin{equation}
N_{q} = \underset{i: 1 \leq i \leq
  n_{train}}{\operatorname{argmin_{\mathcal{K}}}} \sum_{j \in p_v} dist( \underline{h}_{qj}, \underline{h}_{ij} )  
\end{equation}

The above equation simply picks those images that are similar to the
input query image $q$ in the visible parts. If a simple $argmin$ was
defined, it would return the closest neighbor. However, we want the
$\mathcal{K}$ closest neighbors and hence the notation $argmin_\mathcal{K}$.

Once nearest neighbors are picked, the goal is to infer the hidden
parts. For this purpose,  we accumulate representations for hidden
part as $\{\underline{h}_{ij} | i \in N_q, j \in p_h\}$.

We infer the missing part as, $\underline{h}_{qj} = C_j(\underline{h}_{ij}); i \in N_{q}, j\in p_{H}$. Here $C_j: \Re^{|N_q \times K|} \rightarrow \Re^K$ is the consensus function for part $j$. We use the simplest average consensus function:  
\begin{equation}
\{\underline{h}_{ij} | i \in N_q, j \in p_h\} \xmapsto{C_j(\underline{h}_{ij})} \underline{h}_{qj} =
\frac{1}{N_q} \sum_{i \in N_q} \underline{h}_{ij}
\end{equation}

\textbf{Corollary:} \emph{KNN Consensus-Diversity - }
In a generic shopping experience, a shopper would not like to be
presented with the same type of clothing multiple times. As a result,
a ``diverse'' retrieval is required. We propose the following generic optimization for generating diverse
transformed queries, each of which can be used to query the inventory
for similar images. The following optimization is proposed for the
purpose:
\begin{equation}
\mathbf{I}_{q}^{diverse} = Div(\{\underline{h}_{ij} | i \in N_q,  j
\in p_H \})
\label{eqn:diversity}
\end{equation}
$Div$ is an operator that returns a subset $\mathbf{I}_{q}^{diverse}$ of images from the set $N_q$ that are as
different as possible from one another on the hidden part features. In our case $Div$ is a
non-linear operator that clusters the data points and samples points,
one from each cluster.

\subsection{Gaussian Mixture Models (GMM)}
An alternative way of viewing the above problem is using mixture
models. Assuming each dimension of the space in which the Gaussian mixture
model (GMM) is defined to correspond to part features, the GMM intuitively captures the
part features that co-occur most frequently. At test time, the goal is to
efficiently sample these high density regions using features from
visible parts to recommend clothing
on the hidden parts. 

The learning module aims to learn the parameters of a mixture of multivariate Gaussian
distribution parameterized by a random variable $\underline{H}$, and denoted by:
\begin{equation}
p(\underline{H} |\lambda) = \sum_{i=1}^M w_i g( \underline{H}  \mu_i, \mathbf{\Sigma}_i) 
\end{equation}


In the above equation, $g( \underline{H} | \mu_i, \Sigma_i) = \mathcal{N}\left(\mu_i, \mathbf{\Sigma}_i\right)$, $\lambda  = (\underline{\mu}, \mathbf{\Sigma}, \underline{w} )$ denotes the parameters of the Gaussian mixture
distribution having $M$ mixture components, $\underline{\mu} = [\underline{\mu}_1
... \underline{\mu}_M], \mathbf{\Sigma} = [\mathbf{\Sigma}_1 ... \mathbf{\Sigma}_M]$ and $\underline{w} = [w_1 ... w_M]$
respectively denote the means, co-variance matrices and mixture weights of the $M$
Gaussian mixture components. 

Assume query vector $\underline{H}_q^T =
[\underline{h}_{q1}^T, ..., \underline{h}_{qj}^T,   ...,  \underline{h}_{qP}^T]$, with a
  missing part $m$, the goal is to infer the most probable value 
$\underline{h}_{qm}$ of missing part $m$. 

We cast it as the following problem:
\begin{equation}
\hat{\underline{h}}_{qm} = \underset{\underline{h}_{qm}: \underline{H}_q^T =
[\underline{h}_{q1}^T, ..., \underline{h}_{qm}^T,   ...,  h_{qP}]^T}{\operatorname{argmax}} p(\underline{H}|\lambda)
\end{equation}

This is a constrained conditional maximization and will
fix values of parts that have already been observed and will only
search over unknown variables for a maximum likelihood assignment
score.

Since the mixture model defined above requires high dimensional
density estimation, we first vector quantize the part feature space.
Assuming that $P$ parts share a common code book of size, 
a mixture model in $P$ dimensions results.  In other words,
we utilize the code word to which a part feature is most closely
associated with as an input to learn the mixture model.

\subsection{Markov Chain - LDA (MCL)}

\begin{figure}
\centering
\includegraphics[width=0.7\linewidth]{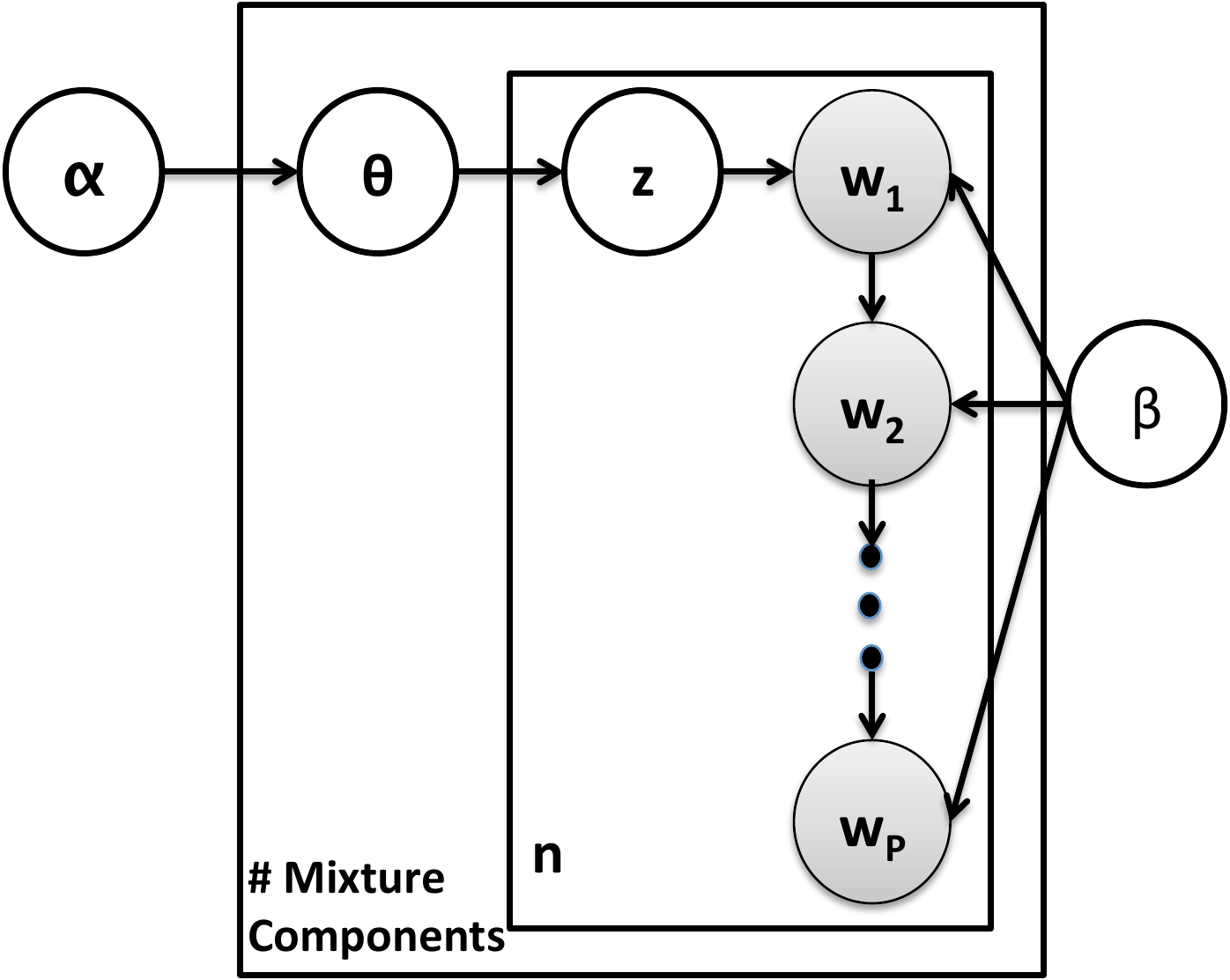} 
\caption{Markov Chain LDA model for learning fashion
  co-occurrences using a topic model, with $w_i$ as the word for part $i$.}
\label{fig:ldaMarkov}
\vspace{-2ex}
\end{figure}

An approach for retrieval using inherent cluster structure of the
data is to utilize topic models. In essence, what we require is a topic
model where words have a specific structure. In other words,
even though words are drawn from a code book, certain word combinations
(say, color combinations) have a much higher probability of occurrence than
others. We propose to model the structure in word combinations by a
Markov Chain (Figure~\ref{fig:ldaMarkov}). The conditional independence properties yielded by a
Markov Chain significantly reduces the computational needs of
structured word generation. 

The generative model for the original LDA
model \cite{BleiLDA03} for words $w$, topics $\theta$ and hyper parameters $\alpha$ and $\beta$ is given by:
\begin{equation}
p(w | \alpha, \beta)  = \int p(\theta | \alpha) \prod_{i=1}^n \sum_{z_i}
p(z_i|\theta) p(w | z_i, \beta) d\theta
\end{equation}


We propose the following generalization of basic LDA to Markov chains
on words, thus modifying the model to:
\begin{equation}
p(\underline{w} | \alpha, \beta)  = \int p(\theta | \alpha) \prod_{i=1}^n \sum_{z_i}
p(z_i|\theta) p(\underline{w} | z_i, \beta) d\theta
\end{equation}
where $\underline{w} = [w_1, w_2,\ldots,w_P]$. Inference over the
joint distribution is computationally expensive. Hence we make the 
following simplifying assumption:
\begin{eqnarray}
p(\underline{w} | \alpha, \beta)  = \int p(\theta | \alpha) \prod_{i=1}^n \sum_{z_i}
p(z_i|\theta) \cdot \nonumber \\   \underbrace{p(w_1) \prod_{j=2}^{P} p(w_j |
w_{j-1}, \beta, z_i )}_{p(\underline{w} | z_i, \beta)}   d\theta
\end{eqnarray}

This modified model is now learned offline using training data. When
a new query comes in, say a shirt, the topic most likely to have this
word is now picked and the Markov chain is sampled to generate a new
transformed sample that can be utilized for retrieving complementary
nearest neighbors.

\subsection{Texture Agnostic Retrieval (TAR)}
We propose the Texture Agnostic Retrieval technique, based on the
notion of consumers preferring \textit{solid} colors to go well with
\textit{patterned} clothing. In other words, having busy patterns in
the top and bottom clothing seems counter-intuitive. The recommender
can be stated as follows: $\underline{h}_{ij} \sim P( \underline{h}, \alpha )$
where $\alpha$ parameterizes the distribution which can be sampled efficiently.
Since the bias has to be towards solid colors, we sample with a
constraint that 
\begin{equation}
\underline{h}_{ij} \sim P( \underline{h}, \alpha ), \text{ s.t. } 
\underline{h}_{ij}^T\mathbf{1}_K = 1
\end{equation}
For our application of interest, we select $P(\underline{h}, \alpha) =
U([0,1])$, while we note any other distribution that can be
efficiently sampled can be employed.

The methods described till now are summarized based on ease of
training, ease of testing, scalability, and generalization in
Table~\ref{table:QuantEvalRecsys}. The requirements for a specific
application would dictate the method of choice for recommendation. We also attempted adopting the collaborative filtering formulation of~\cite{RuslanPMF08}, but do not describe it in detail due to practical difficulties we encountered in scaling the algorithm to large data sets.

\section{Experiments}
\label{sec:Experiments}
Our experiments are conducted on the data sets summarized in Table~\ref{table:Datasets}.

\subsection{Low-Level Representation}
\label{sec:LowLevel}
We experiment with variations of color based features in the model though the proposed formulations can very easily be extended to other descriptor based features such as Dense-SIFT~\cite{BoschECCV06} as well. We experiment with two low-level representations of image, namely (i) HSV Histograms: Given $P$ parts of interest characterized by $P$ distinct bounding boxes, a normalized  $40$ dimensional color histogram is computed by quantizing Hue Saturation and Value into $24,8,8$ uniformly spaced bins respectively, (ii) Color BoW:  A $K$-dimensional feature vector is obtained by randomly sampling fixed size ($15 \times 15$) patches from the bounding box and quantizing them to a learned code book of $K$ code book entries.

\begin{table}[t]
\caption{Comparison of different models: + (Easy), o (Medium), x (Hard)}
\begin{tabular}{p{.6\linewidth}p{.3\linewidth}}
{
{
	\footnotesize
	{
		\begin{tabular}{p{1.6cm} | p{0.6cm} p{0.5cm} p{0.5cm} p{0.5cm} p{0.5cm}}
		\hline
		\textbf{Model:} & \textbf{CNNC}  &  \textbf{GMM}  &  \textbf{TAR} &  \textbf{MCL} & \textbf{PR} \\
		\hline
		\textbf{Learn Ease}  & +  &  +  &  + &  x  & + \\
		\textbf{Test Ease}  & x  &  x  &  + &  + & + \\
		\textbf{Scalability} &  o  &  +  &  + &  o & +\\
		\textbf{Generalization}  & +  &  o  &  o &  + & x \\
		\hline
		\end{tabular}
	}
}
}
\end{tabular}
\vspace{-0.8em}
\label{table:QuantEvalRecsys}
\end{table}
%
%

\begin{table}[t!]
\centering
\caption{Summary of fashion data sets in our study.}
\footnotesize
{
\begin{tabular}{lp{5.5cm}}
\hline
\textbf{Data set} & \textbf{Comments} \\
\hline\hline \\ [-2ex]
\textbf{Fashion-136K} &  135893 street fashion images with annotations by fashionistas, brand, demographics.\\
\textbf{Fashion-Toy} & Subset of Fashion-136K. $n=600$, $n_{train}=500$, $n_{test}=100$. \\
\textbf{Fashion-63K} & Subset of Fashion-136K. $n=63K$, $n_{train}=53K$, $n_{test}=10K$\\
\textbf{Fashion-350K} & 350K images of just tops \& blouses to test retrieval performance using fashionistas.\\
\textbf{Fashion-Q1K} & 1K images of skirts used to retrieve images from Fashion-350K. Skirts have one of the following different patterns: animal-print (100), floral (200), geometric (100), plaids \& checks (150), paisley (50), polka dots (100), solid (200), stripes (100). \\
[1ex]
\hline
\end{tabular}
}
\vspace{-2ex}
\label{table:Datasets}
\end{table}

We performed extensive experiments with texture features such Gabor Wavelets~\cite{MaCVPR96}, Textons~\cite{varma2005statistical},
Dense-SIFT\cite{BoschECCV06} and found them to be unstable for texture classification in clothing. A preliminary analysis suggests that deformable nature of clothing often leads to unreliable texture estimation where even wrinkles and folds on solid clothes can be misinterpreted as texture. Hence, in this paper we focus our attention to using only color based image representations.

\subsection{Experimental Setup for Retrieval} In the past sections, we presented methods to learn the missing part of a query image. For instance, given a descriptor for skirt, what is the recommended descriptor for blouse? Thus the query descriptor is essentially mapped to a new space (say, from skirt to blouse). This mapped query now serves as an input query to a content based image retrieval (CBIR) system, for retrieving images similar to the mapped query. 

\textit{\textbf{Training:}} The Fashion-136K is a data set created by crawling the web for photographs of fashion models. Hence, all images comprise top and bottom clothing co-occurring in the same image. Further, their spatial location $(x,y)$ coordinates are also manually annotated by fashionistas. We use this data set for training all data-driven models described previously. While studying the data, we observed interesting co-occurrence relationships between top and bottom clothing such as dark gray/black bottoms going well with any colored top. It is such interesting patterns that the data driven models attempt to learn. In scenarios that require recommendations to be driven by context (Figure~\ref{fig:themeQuery}), it is possible to learn a separate co-occurrence matrix for every context.

\textit{\textbf{Retrieval:}} Images of skirts (bottom clothing) from the Fashion-Q1K dataset are utilized as queries to retrieve top clothing from the Fashion-350K data set. Fashion-350K images are from a clothing inventory containing only top clothing (without model or mannequin). As a result, we utilize Fashion-136K where top and bottom clothing \emph{co-occur} to learn models that are used while querying Fashion-350K. It is useful to note that this procedure trains the retrieval system on images on a data set (Fashion-136K) that is completely different from the test data (Fashion-350K). The result of retrieval is a ranked list of the Fashion-350K images sorted by relevance to a query from Fashion-Q1K. Since there are five algorithms proposed in this work, the results of retrieval are five ranked lists where each list corresponds to a different algorithm's output.

\begin{figure*}
\centering
\begin{tabular}{ccc}
\includegraphics[width=.3\linewidth]{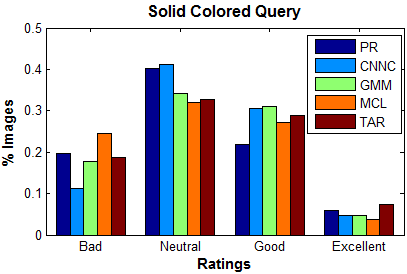}  &
\includegraphics[width=.3\linewidth]{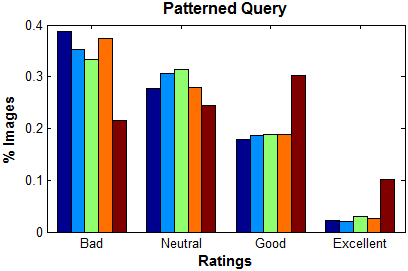}  &
\includegraphics[width=.3\linewidth]{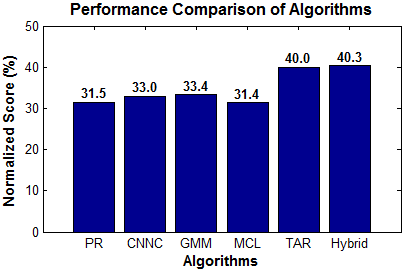} \\
(a) & (b) & (c)
\end{tabular}
\caption{Performance comparison of algorithms. CNNC performs the best for solid colored queries (20\% of query set Fashion-Q1K) and TAR for patterned queries. (a) Ratings by fashionistas for solid colored queries. (b) Ratings for patterned queries. (c) Overall performance for the complete query set (includes both solid colors and patterns). The hybrid approach uses  CNNC for solids and TAR for patterned queries, based on the decision of a solid vs pattern classifier.}
 \label{fig:amtQuantPlots}
\vspace{-2ex}
\end{figure*}

\subsubsection{Experimental Validation} In order to validate the results of retrieval, we utilize fashionistas. Since each query has five ranked lists as the output of retrieval, we retain the top ten relevant results from each algorithm to measure retrieval quality. Each fashionista is presented with a query image, and an image grid having 10 columns corresponding to top ten retrievals, and 5 rows corresponding to each algorithm.  We present these results to fashionistas in random order (to offset presentation bias) and ask them to rate each algorithm overall as one of
-1 (bad match), 0 (neutral match), 1 (good match), 2 (excellent match).
This validation procedure is in line with the established validation protocols for image retrieval~\cite{yao2011classifying}. All thousand queries were validated by five fashionistas per query, leading to a total of
5000 responses. However, since fashion tastes vary greatly across people, it is usually hard to get a consensus on which algorithm performs the best from all 5 fashionistas. As a result, we only retain ratings from fashionistas who \textit{agree} on algorithm performance.

Denoting each rating of 5 algorithms (PR, CNNC, GMM, MCL, TAR) for a query $q$ using a five dimensional vector $\underline{\zeta}_{qi} \in \{-1, 0, 1, 2\}^5$, where each entry is a rating from -1 to +2, the score for disagreement across fashionistas is defined as, $\gamma_{qi} = \sum_{j=1}^5 ||\underline{\zeta}_{qi} - \underline{\zeta}_{qj}||_{1}$.

Further, an agreement threshold is defined to be the median of disagreement scores across all queries, $A_T = median(\gamma_{qi});
q \in [1, 1000], i \in [1, 5]$. Fashionista ratings are retained only if $\gamma_{qi} < A_T$. Finally, a query rating confidence is computed as $C_q = \frac{\text{Number of Fashionistas Retained for a Query}}{\text{Total Number of
  Fashionistas for a Query}}$. The final score $S_a$ for an algorithm $a$ derived from fashionistas is computed as:
\begin{equation}
S_a = \sum_{q=1}^{1000} C_q \frac{ \sum_{i=1}^5 \zeta_{qi} (a) \delta(
  \gamma_{qi} < A_T) }{\sum_{i=1}^5 \delta( \gamma_{qi} < A_T)} \end{equation} where $\delta(x)$ is the Dirac-delta function, $\zeta_{qi} (a)$ refers to the rating provided by fashionista $i$ on query $q$ for algorithm $a$. Since there are inadvertent errors by fashionistas, the total number of ratings that result after a preliminary filtering of ratings was 937, split across 187 solid queries and 750 patterned queries.

\subsubsection{Analysis of Results}
\begin{figure*}[t!]
\renewcommand\arraystretch{1.2} \addtolength{\minrowclearance}{3pt}
\centering

\begin{tabularx}{\linewidth}{c|c|c|c|c|c|c}
\footnotesize{\bf{Type}}& \footnotesize{\bf{Query}} & \footnotesize{\bf{PR}} & \footnotesize{\bf{CNNC}} & \footnotesize{\bf{GMM}} & \footnotesize{\bf{MCL}} & \footnotesize{\bf{TAR}}\\
\toprule

\begin{sideways}\begin{minipage}[t]{0.5cm} \footnotesize{\bf{Animal Print}}\end{minipage}\end{sideways} &
\includegraphics[width=0.065\linewidth]{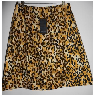} &
\includegraphics[width=0.145\linewidth]{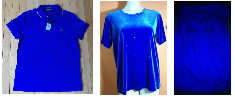}     &
\includegraphics[width=0.145\linewidth]{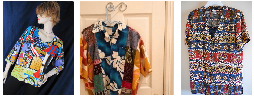}   &
\includegraphics[width=0.145\linewidth]{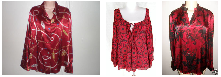}    &
\includegraphics[width=0.145\linewidth]{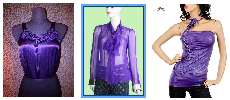}    &
\includegraphics[width=0.145\linewidth]{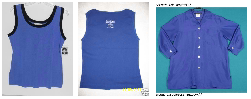} 
\\
\hline
\begin{sideways}\footnotesize{\bf{Floral}}\end{sideways} &
\includegraphics[width=0.05\linewidth]{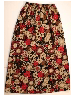} &
\includegraphics[width=0.145\linewidth]{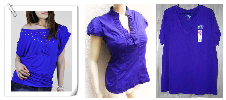}   &
\includegraphics[width=0.145\linewidth]{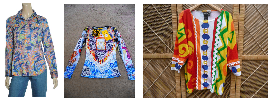}  &
\includegraphics[width=0.145\linewidth]{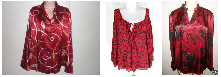}   &
\includegraphics[width=0.145\linewidth]{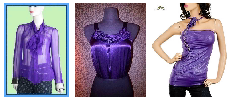}  &
\includegraphics[width=0.145\linewidth]{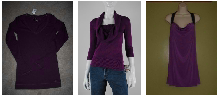} 
\\
\hline
\begin{sideways}\footnotesize{\bf{Geometric}}\end{sideways} &
\includegraphics[width=0.07\linewidth]{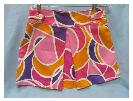} &
\includegraphics[width=0.145\linewidth]{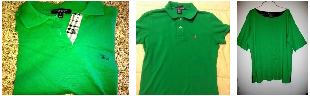}    &
\includegraphics[width=0.145\linewidth]{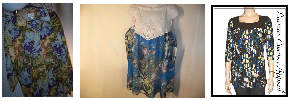}  &
\includegraphics[width=0.145\linewidth]{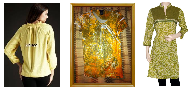}   &
\includegraphics[width=0.145\linewidth]{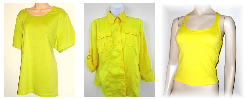}   &
\includegraphics[width=0.145\linewidth]{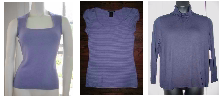}
\\
\hline
\begin{sideways}\footnotesize{\bf{Paisley}}\end{sideways} &
\includegraphics[width=0.04\linewidth]{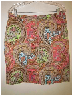} &
\includegraphics[width=0.145\linewidth]{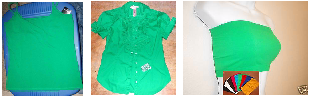}     &
\includegraphics[width=0.145\linewidth]{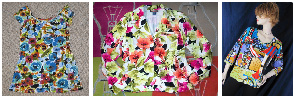}   &
\includegraphics[width=0.145\linewidth]{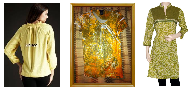}    &
\includegraphics[width=0.145\linewidth]{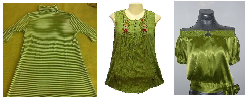}    &
\includegraphics[width=0.145\linewidth]{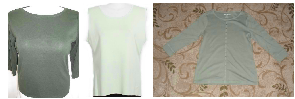}    
\\
\hline
\begin{sideways}\footnotesize{\bf{Plaids}}\end{sideways} &
\includegraphics[width=0.05\linewidth]{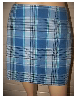} &
\includegraphics[width=0.145\linewidth]{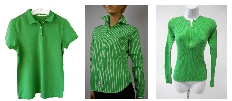}     &
\includegraphics[width=0.145\linewidth]{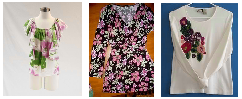}   &
\includegraphics[width=0.145\linewidth]{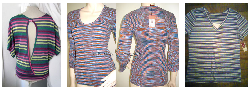}    &
\includegraphics[width=0.145\linewidth]{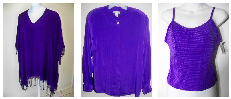}    &
\includegraphics[width=0.145\linewidth]{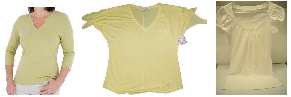}
\\
\hline
\begin{sideways}\begin{minipage}[t]{0.5cm} \footnotesize{\bf{Polka Dots}}\end{minipage}\end{sideways} &
\includegraphics[width=0.06\linewidth]{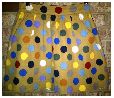} &
\includegraphics[width=0.145\linewidth]{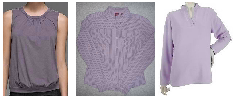}   &
\includegraphics[width=0.145\linewidth]{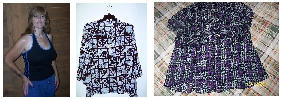}  &
\includegraphics[width=0.145\linewidth]{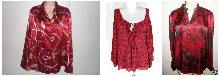}   &
\includegraphics[width=0.145\linewidth]{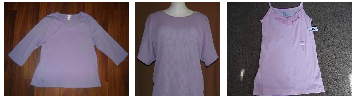}   &
\includegraphics[width=0.145\linewidth]{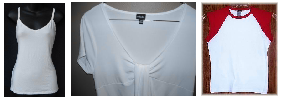} 
\\
\hline
\begin{sideways}\footnotesize{\bf{Solids}}\end{sideways} &
\includegraphics[width=0.054\linewidth]{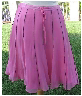} &
\includegraphics[width=0.145\linewidth]{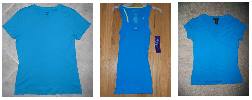}     &
\includegraphics[width=0.145\linewidth]{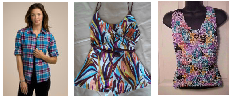}   &
\includegraphics[width=0.145\linewidth]{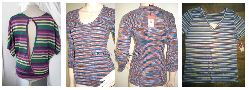}    &
\includegraphics[width=0.145\linewidth]{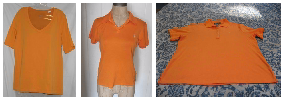}    &
\includegraphics[width=0.145\linewidth]{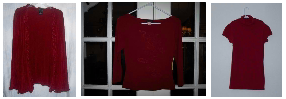} 
\\
\hline
\begin{sideways}\footnotesize{\bf{Striped}}\end{sideways} &
\includegraphics[width=0.045\linewidth]{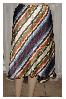} &
\includegraphics[width=0.145\linewidth]{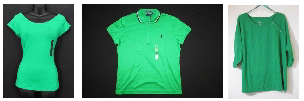}    &
\includegraphics[width=0.145\linewidth]{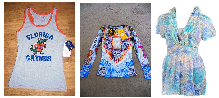}  &
\includegraphics[width=0.145\linewidth]{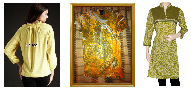}   &
\includegraphics[width=0.145\linewidth]{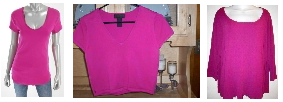}   &
\includegraphics[width=0.145\linewidth]{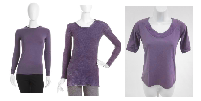} 
\\
\bottomrule
\end{tabularx}
\vspace{0.5em}
\caption{Top 3 retrieved items from ``tops", recommended by each algorithm for query ``skirts". Rows correspond to a query from a given pattern. Recommendations by TAR are more preferred than those by PR and MCL. Recommendations by CNNC and GMM have more patterns, even for patterned queries. Note that code words and modes in GMM yield identical retrievals for multiple queries. }
\label{fig:CompareCBIR}
\end{figure*}

As shown in results from Figure~\ref{fig:amtQuantPlots}(a-c), data driven models outperform perceptual retrieval. We study the performance of algorithms on solid colored and patterned clothing separately to gain better insight into the workings of each algorithm. Further, it is useful to note that we had initially given the fashionistas an option to rate on a scale from Bad-Excellent (-1 to
+2). Visual results of the various retrieval algorithms are presented
in Figure~\ref{fig:CompareCBIR}.

\begin{figure}[h!]
\centering
\vspace{-1ex}
\includegraphics[width=.8\linewidth]{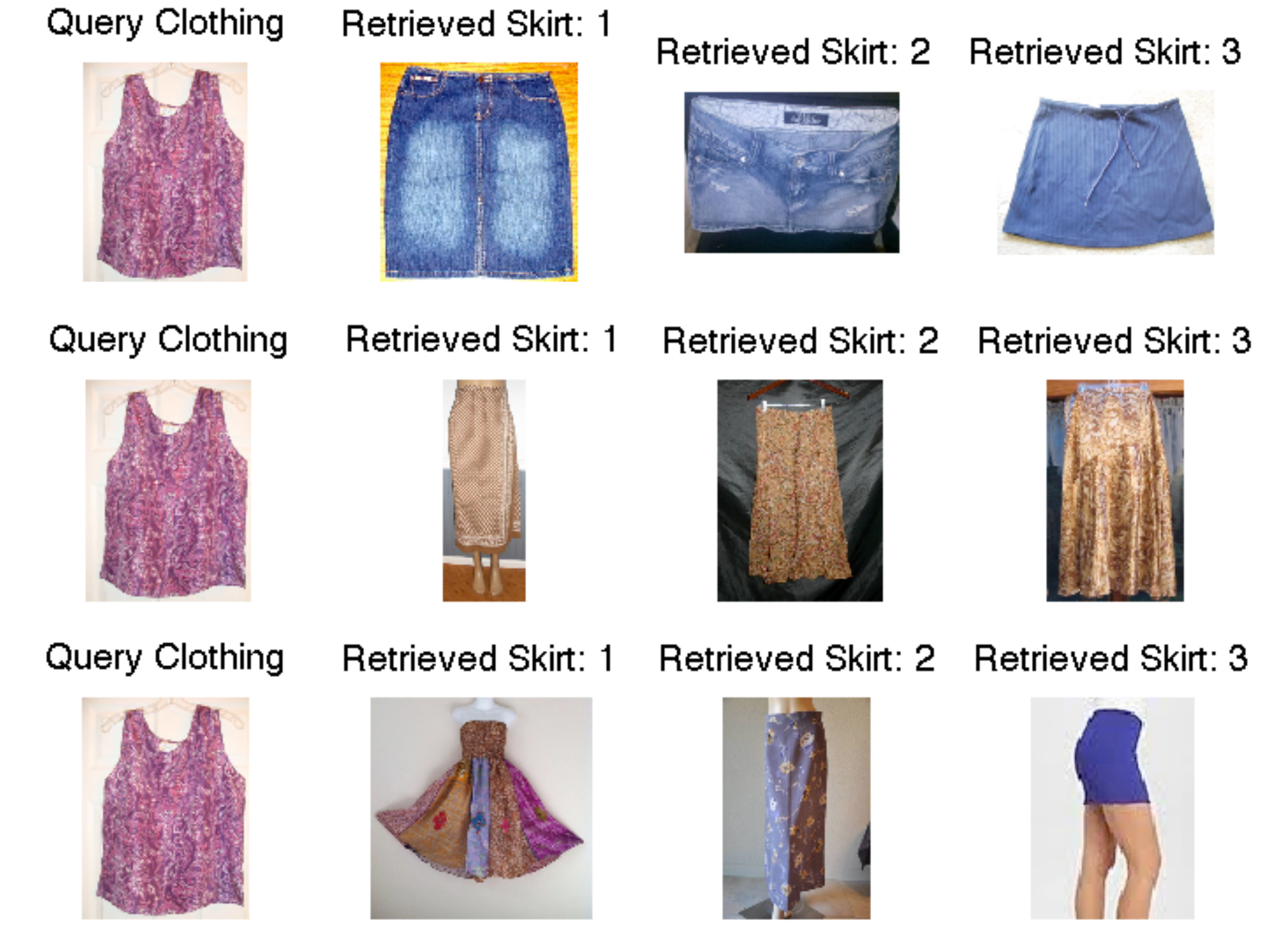}  
\caption{Illustration of diverse retrievals using CNNC, corresponding
to a sample query. Each row corresponds to different set of results, while being relevant to the query.}
\label{fig:diversity}
\vspace{-0.5ex}
\end{figure}

\textbf{\textit{Retrieval for Solid Colored Query:}} On solid clothing queries, it was observed that CNNC had the best normalized score of $0.418$ on a scale from $0$ to $1$, followed by TAR $0.402$, GMM $0.398$, PR $0.383$ and MCL $0.370$. As seen in Figure~\ref{fig:CompareCBIR}, CNNC tends to retrieve patterned clothing. This is in fact favored by fashionistas in our study, for solid colored queries. The strength of adding diversity to a method like CNNC, as in Equation~\ref{eqn:diversity}, is also shown qualitatively in Figure~\ref{fig:diversity}.

\textbf{\textit{Retrieval for Patterned Query:}} On pattern clothing queries, TAR performed the best with a normalized score of $0.40$ on a scale from $0$ to $1$, followed by GMM $0.31$, CNNC $0.30$, MCL $0.30$ and PR $0.29$. The impressive performance of TAR can be attributed to the general preference for solid colored clothing to go with another patterned clothing. Recall that by construction TAR is agnostic to texture and recommends only solid colors. Further, the simplicity of TAR leads to a fairly robust query transformation. Finally, the query for solid colors by TAR leads to a lot more relevant retrievals from Fashion-350K, since it is much easier to match and retrieve solid clothing from an inventory in comparison to patterned clothing.

\textbf{\textit{The Hybrid Classifier:}} Based on our findings, we observe that CNNC performs the best for solid queries, while TAR performs best on pattern queries. Since a query could either be a solid or patterned, we propose using a single hybrid recommender that resorts to using CNNC for solid queries and TAR for patterned queries. The switching can be determined at the beginning of the retrieval procedure using a simple pattern classifier that labels a query as either solid or patterned. Experimental results on the overall query set (solid+patterns) indicate that the hybrid classifier which switches between CNNC and TAR yields the highest overall performance of $0.403$, see Figure~\ref{fig:amtQuantPlots}(c).

\begin{figure*}[t!]
\centering
\begin{tabular}{ccc}
\includegraphics[width=.3\linewidth]{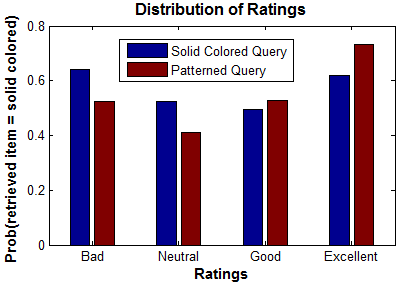}  &
\includegraphics[width=.3\linewidth]{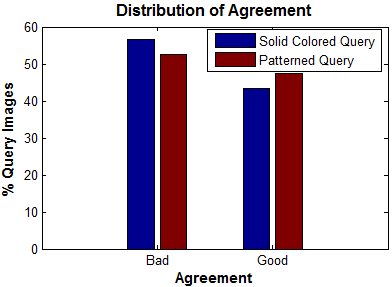}  &
\includegraphics[width=.3\linewidth]{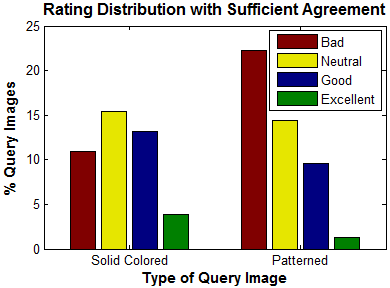} \\
(a) & (b) & (c) 
\end{tabular}  
\caption{(a) A solid vs pattern classifier was used to estimate probability of solid colored clothing in the top retrievals. Fashionistas prefer patterned recommendations for solid colored queries and vice versa. (b) Illustration of rating agreement on solid colored and  patterned queries. Fashionistas tend to agree  more on retrievals of patterned queries, than they agree on solid queries. (c) This expands the second half of (b). Depicts common agreement amongst fashionstas based on retrieved results. Results for solid colored queries are generally more favorable than for patterned queries.}  \label{fig:auxExperiments}
\vspace{-2ex}
\end{figure*}

\textbf{\textit{Ratings for Solid Colored Retrievals:}} The previous quantitative experiments seem to offer an insight that fashionistas prefer solid
retrievals for patterned queries (evidenced by the impressive
performance of TAR), which we validate in this experiment by exploring how fashionistas respond to solid colored retrievals. In other words, we run a binary classifier on the retrieval set to obtain the average probability of the retrieval set to be solid. When considering solid queries (denoted by blue bars) in Figure~\ref{fig:auxExperiments}(a), we infer that fashionistas do not favor solid colored retrievals. This is evidenced by the higher magnitude of the blue bars on \textit{Bad} and \textit{Neutral} ratings. On the other hand, when considering patterned queries (denoted by red bars) in Figure~\ref{fig:auxExperiments}(a), we infer that fashionistas tend to favor solid colored retrievals. Observe the higher magnitude of the red bars on \textit{Good} and \textit{Excellent} ratings. This provides an intuitive justification behind the impressive performance of TAR on the pattern queries, since solids are retrieved by construction.

\textbf{\textit{Distribution of Rating Agreement:}} Next, we study the rating agreement on solid and patterned queries. In other words, we measure how \textit{agreeable} the recommendation algorithm's results are across solid and patterned queries, see Figure~\ref{fig:auxExperiments}(b). It can be readily observed that fashionistas tend to agree more on patterned queries, evidenced by the higher magnitude of red bars on the good agreement bin. On the other hand, fashionistas tend to disagree more on the solid queries in comparison to patterns. It would be interesting to study whether rating agreement was on retrievals that were rated as \emph{Good}, or on retrievals they thought were \emph{Bad}. Figure~\ref{fig:auxExperiments}(c) shows the split of among the ratings that fashionistas agreed on. In the case of solids, the fashionistas agreed that the retrieval was either \emph{Neutral} or \emph{Good} (evidenced by the higher weights to the yellow and blue bars). However, on patterned queries, the fashionistas overwhelmingly agreed that the retrievals were not favorable (evidenced by the higher weights on the red bar). This finding indicates that the performance of the algorithms on pattern queries can be further enhanced. A major reason for this result is the fact that the proposed system is
purely color based, and integration of novel texture features devoid
of drawbacks mentioned in Section~\ref{sec:LowLevel} could possibly yield a significant performance boost. This is a promising avenue for future research.

\textbf{\textit{Insights into Street Fashion:}} Our experimental analysis leads us to the following insights on street fashion, (i) Fashion Cues:
Analyzing fashionista ratings for different algorithms suggest an intuitive fashion cue that a pair of patterned and solid colored clothing is more perceptually pleasing than other combinations. It also suggests that even among matches for patterned clothing, color is more important than the type of texture (i.e. plaids, polka dots), which also underscores the importance of simple yet visually strong descriptors such as color, (ii) Among all the patterns, we observed that fashionistas agree the most on recommending matches for paisley and stripes. However, fashionistas have more well-defined preferences for stripes as compared to paisley. Our initial analysis suggests this behavior can be attributed to the structure of pattern:
stripes have strong structural information as compared to paisley. Hence, it is easier to recommend matches for stripes than for paisley.

\section{Conclusion and Future Work}

In this paper, we presented data-driven strategies to understand street fashion. We proposed 4 models for this purpose and conducted extensive experiments to illustrate their efficacy over perceptual models. Our main finding from this work is the power of data-driven analysis over perceptually driven models. This is  attributed to the fact that data-driven models adapt well to the given domain of fashion as opposed to perceptually driven models that are very generic. Moreover, these techniques allow us to learn intuitive domain insights directly from the data, without requiring human in the loop. 

Our future work focuses on exploring different combination methods to leverage relative strengths of various models. We also aim to study the perceptual bias present in fashionistas preferences and use it to design better personalized experiences for fashion related shopping. Further, there are key fashion insights present in large image repositories derived from fashion blogs, that can be successfully exploited for a number of applications. Such tagged data sets like Fashion-136K (see Figure~\ref{fig:AvgImageCorpus}) can also be used for fundamental vision research problems such as human pose estimation, semantic segmentation and object localization.

\definecolor{Gray}{gray}{0.87}
\definecolor{LightGray}{gray}{0.95}
\newcolumntype{g}{>{\columncolor{Gray}}c}
\newcolumntype{w}{>{\columncolor{LightGray}}c}

{\small
\bibliographystyle{ieee}
\bibliography{bib}
}

\end{document}